%% file: main.tex
\documentclass[acmtog]{acmart}
\author{Beijia Lu}
\email{beijialu@andrew.cmu.edu}
\affiliation{
  \institution{Carnegie Mellon University}
  \country{U.S.A}
}

\author{Ziyi Chen}
\email{ziyi.paii@gmail.com}
\affiliation{
  \institution{PAII Inc.}
  \country{U.S.A}
}

\author{Jing Xiao}
\email{xiaojing661@pingan.com.cn}
\affiliation{
  \institution{PAII Inc.}
  \country{U.S.A}
}

\author{Jun-Yan Zhu}
\email{junyanz@andrew.cmu.edu}
\affiliation{
  \institution{Carnegie Mellon University}
  \country{U.S.A}
}

\AtBeginDocument{%
  }

\input{packages}
\input{sections/macros}
\copyrightyear{2025}
\acmYear{2025}
\setcopyright{cc}
\setcctype{by}
\acmConference[SA Conference Papers '25]{SIGGRAPH Asia 2025 Conference Papers}{December 15--18, 2025}{Hong Kong, Hong Kong}
\acmBooktitle{SIGGRAPH Asia 2025 Conference Papers (SA Conference Papers '25), December 15--18, 2025, Hong Kong, Hong Kong}\acmDOI{10.1145/3757377.3763831}
\acmISBN{979-8-4007-2137-3/2025/12}
\begin{document}
\title{Input-Aware Sparse Attention for Real-Time Co-Speech Video Generation}
\renewcommand{\shortauthors}{Lu, Chen, Xiao, and Zhu}

\begin{abstract}
Diffusion models can synthesize realistic co-speech video from audio for various applications, such as video creation and virtual agents. However, existing diffusion-based methods are slow due to numerous denoising steps and costly attention mechanisms, preventing real-time deployment. In this work, we distill a many-step diffusion video model into a few-step student model. Unfortunately, directly applying recent diffusion distillation methods degrades video quality and falls short of real-time performance. 
To address these issues, our new video distillation method leverages input human pose conditioning for both attention and loss functions. We first propose using accurate correspondence between input human pose keypoints to guide attention to relevant regions, such as the speaker's face, hands, and upper body. This input-aware sparse attention reduces redundant computations and strengthens temporal correspondences of body parts, improving inference efficiency and motion coherence. To further enhance visual quality, we introduce an input-aware distillation loss that improves lip synchronization and hand motion realism. By integrating our input-aware sparse attention and distillation loss, our method achieves real-time performance with improved visual quality compared to recent audio-driven and input-driven methods. We also conduct extensive experiments showing the effectiveness of our algorithmic design choices. 
\end{abstract}


\begin{teaserfigure}
  \includegraphics[width=\textwidth]{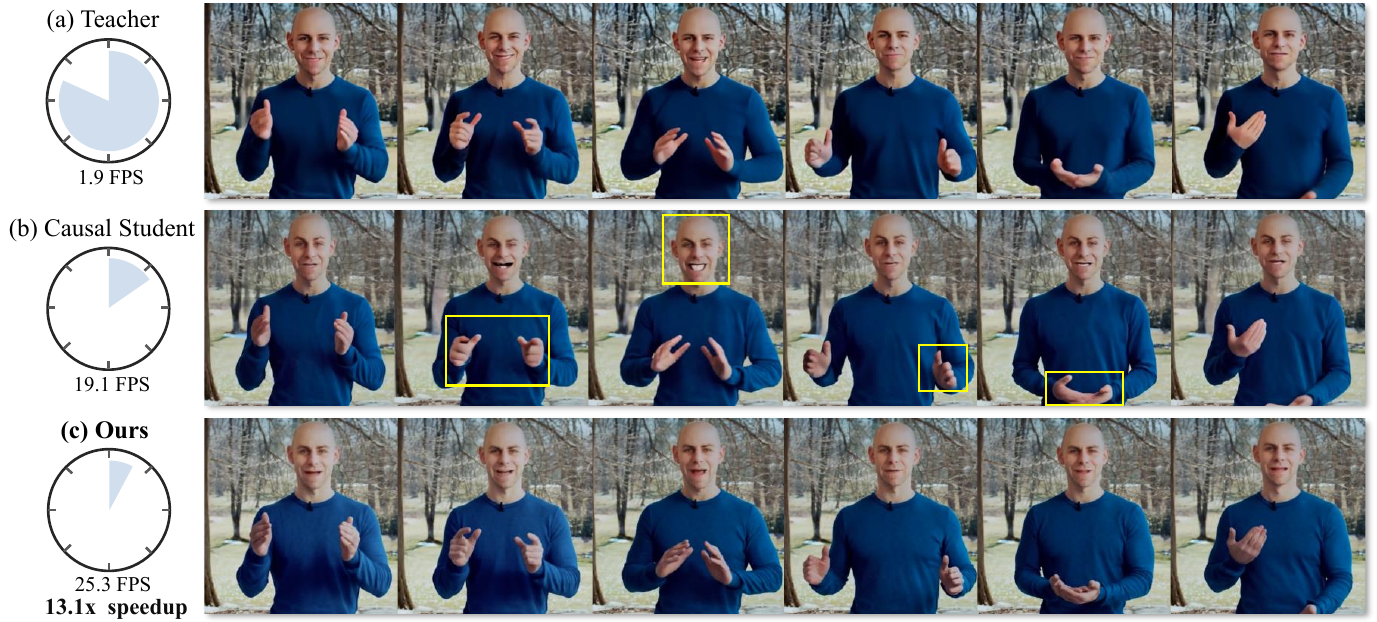}
  \caption{We introduce a conditional video distillation method for real-time co-speech video generation that leverages human pose conditioning for input-aware sparse attention and distillation loss. Our student model achieves 25.3 FPS, a $13.1\times$ speedup over its teacher model~\cite{zhang2024mimicmotion}, while preserving visual quality. Our method significantly improves motion coherence and lip synchronization over a leading few-step causal student model~\cite{yin2025causvid}, while reducing common visual degradation in the speaker's face and hands (see yellow box). }
  \label{fig:teaser}
\end{teaserfigure}

\maketitle
\input{sections/1_intro}

\input{sections/2_related_work}

\input{sections/3_method}

\input{sections/4_result}

\input{sections/5_conclusion}

\bibliographystyle{ACM-Reference-Format}
\bibliography{main}

\end{document}

%% file: packages.tex
\usepackage[legacy]{csvsimple}
\usepackage{multirow}
\usepackage{makecell}
\usepackage{arydshln}
\usepackage{tabularx}

\usepackage{graphicx}
\usepackage{duckuments}

\usepackage{subcaption}
\usepackage{caption}
\usepackage{enumitem}

\usepackage{float}

%% file: sections/macros.tex
\def\x{{\mathbf{x}}}

\makeatletter
\newcommand{\smallsym}[2]{#1{\mathpalette\make@small@sym{#2}}}
\newcommand{\make@small@sym}[2]{%
  \vcenter{\hbox{$\m@th\downgrade@style#1#2$}}%
}
\newcommand{\downgrade@style}[1]{%
  \ifx#1\displaystyle\scriptstyle\else
    \ifx#1\textstyle\scriptstyle\else
      \scriptscriptstyle
  \fi\fi
}
\makeatother

\newcommand{\reffig}[1]{Figure~\ref{fig:#1}}
\newcommand{\refsec}[1]{Section~\ref{sec:#1}}

\newcommand{\lblfig}[1]{\label{fig:#1}}
\newcommand{\lblsec}[1]{\label{sec:#1}}

\newcommand{\ignorethis}[1]{}

\def\1{\bm{1}}

\newcolumntype{L}[1]{>{\raggedright\let\newline\\\arraybackslash\hspace{0pt}}m{#1}}
\newcolumntype{C}[1]{>{\centering\let\newline\\\arraybackslash\hspace{0pt}}m{#1}}
\newcolumntype{R}[1]{>{\raggedleft\let\newline\\\arraybackslash\hspace{0pt}}m{#1}}

\newcommand{\ignore}[1]{}

\makeatletter
\DeclareRobustCommand\onedot{\futurelet\@let@token\@onedot}
\def\@onedot{\ifx\@let@token.\else.\null\fi\xspace}

\makeatother

\definecolor{MyPurple}{RGB}{111,0,255}

%% file: sections/1_intro.tex
\section{Introduction}
\lblsec{intro}
Co-speech video generation, which synthesizes human videos from audio, has enabled diverse applications such as creating realistic virtual agents, producing educational video content, and enhancing telepresence. Diffusion models~\cite{ho2020ddpm}, with attention mechanisms, excel at this task, producing highly realistic and coherent videos. However, diffusion models are extremely slow for video generation, due to the numerous denoising steps and the quadratic costs associated with full attention over many frames and tokens, making real-time deployment difficult.

One common strategy to reduce computational costs is network distillation~\cite{hinton2015distilling}, where a slow, pre-trained ``teacher'' video model is distilled into a faster ``student'' model. Unfortunately, as \reffig{teaser}b shows, directly applying recent text-to-video distillation methods, such as CausVid~\cite{yin2025causvid}, to the co-speech generation task proves insufficient. Such methods tend to degrade the video quality, particularly in crucial areas like the speaker's face and hand regions, which are vital for natural co-speech gestures. Moreover, the speedup achieved still falls short of the demands for real-time performance.

To address these limitations, we propose a conditional video distillation method for fast co-speech generation with the following key insight. Unlike general text-to-video distillation, co-speech generation benefits from readily available input human pose conditioning. We leverage this pose information through an input-aware sparse attention mechanism, which significantly reduces computation while preserving video quality. Pose keypoints allow us to identify key regions (e.g., the speaker’s face, hands, upper body). This enables our sparse attention to selectively focus on tokens within these regions and their corresponding areas in similar frames, rather than performing dense, costly full attention across the entire video frames.
Beyond guiding attention, pose information also pinpoints critical regions, such as the face and hands. This also enables us to design distillation losses that prioritize perceptual quality and accuracy in these areas, crucial for convincing co-speech video. Building on these insights, our method incorporates input-aware sparse attention and distillation loss into an efficient video distillation method. The sparse attention mechanism focuses computation on pose-defined relevant regions, which reduces redundancy and improves motion coherence. The input-aware distillation loss further enhances visual quality, particularly lip synchronization and hand motion realism.

We have conducted extensive experiments on both the publicly available TalkShow dataset~\cite{yi2023generating} and a more comprehensive, newly curated dataset. On both datasets, our method demonstrates significant improvements. It achieves real-time performance, with a $13.1\times$ speedup compared to the teacher model, while concurrently improving upon the teacher model’s visual quality by $10\%$ in lip synchronization and 2\% in motion coherence. Our approach consistently outperforms recent audio-driven and pose-driven methods in terms of generation quality, lip synchronization, and hand motion realism, while being significantly faster. We further include a thorough ablation study to quantitatively and qualitatively analyze the effectiveness of both sparse attention and distillation loss.  To our knowledge, our work presents the first real-time diffusion-based co-speech avatar. Our code and models are available at \url{https://beijia11.github.io/IASA}.

%% file: sections/2_related_work.tex
\section{Related Work}

\paragraph{Co-Speech Video Generation}
Co-speech video generation aims to synthesize human videos driven by audio inputs, ensuring temporal synchronization between speech and human motions, such as facial expressions and hand gestures. One common approach divides this task into two subtasks: generating human motions from speech, and subsequently creating videos from these synthesized motions. Some works solely focus on one of these subtasks~\cite{yi2023generating, liu2023emage,chen2024diffsheg, tian2024emo2,zhang2024mimicmotion}, while others address the task as a whole~\cite{ginosar2019learning,hogue2024diffted,meng2024echomimicv2,Mahapatra2024CoSpeech3D,corona2024vlogger,liu2024tango}. 

Regarding audio-to-motion generation, Speech2Gesture~\cite{ginosar2019learning} uses Generative Adversarial Networks (GANs)~\cite{goodfellow2014generative} to generate human 2D keypoints. TalkShow ~\cite{yi2023generating} and EMAGE~\cite{liu2023emage} use VQ-VAEs~\cite{van2017neural} for generating human 3D meshes, while DiffSheg~\cite{chen2024diffsheg} replaces VQ-VAE with diffusion models~\cite{ho2020ddpm,song2021ddim}. Recent methods~\cite{liu2022audio, he2024co, hogue2024diffted} focus on generating unsupervised motion representations~\cite{siarohin2019first,siarohin2019animating} to further improve the output fidelity. 
In our work, we reuse the EMAGE module~\cite{liu2023emage} and focus on accelerating the motion-to-video part, as audio-to-motion accounts for a small portion of the overall computational costs. 

For the motion-to-video task, conditional GANs~\cite{isola2017image,wang2018vid2vid} have been widely used to inject various motion representations into videos: Speech2Gesture~\cite{ginosar2019learning} uses 2D keypoints, \citet{Mahapatra2024CoSpeech3D} use 3D textured human meshes, and methods like ANGIE~\cite{liu2022audio} and S2G-MDDiffusion~\cite{he2024co} use implicit motion representations~\cite{wang2021one}. However, GAN-based methods often struggled with generalization, tending to overfit to specific identities seen during training. To improve generalization across unseen speakers and motions, recent works have increasingly adopted diffusion models. MM-Diffusion~\cite{ruan2022mmdiffusion} uses a joint diffusion architecture to model multimodal correlations. Other approaches~\cite{guan2024talk,tian2024emo2,zhang2024mimicmotion,meng2024echomimicv2}
have introduced explicit intermediate pose representations as conditioning signals within the diffusion framework. While diffusion-based methods outperform GANs in quality and generalization, their iterative denoising process and costly attention mechanisms hinder real-time deployment. Our work aims to significantly accelerate diffusion-based co-speech avatars.

\paragraph{Diffusion Model Distillation.} 
Diffusion models~\cite{ho2020ddpm,sohl2015deep,song2021ddim} require a large number of denoising steps to achieve high-quality generation, limiting their applicability in low-latency applications. To accelerate inference, a range of distillation techniques have been proposed to reduce the number of steps while preserving visual quality. Trajectory-preserving methods aim to approximate the denoising trajectory of a teacher model~\cite{salimans2022progressive,luhman2021knowledge}. For example, \citet{luhman2021knowledge} train a single-step student to match the teacher's output at the final step, while Consistency Distillation~\cite{kim2024consistency} trains the student to map any points on an ODE trajectory to its origin. Rectified flow ~\cite{liu2024instaflow} trains a student model on the linear interpolation path of noise-image pairs obtained from the teacher. 

In contrast to trajectory-based methods, Distribution Matching Distillation (DMD)~\cite{yin2024improveddmd,yin2024onestepdmd} and adversarial distillation~\cite{sauer2024fast,sauer2024adversarial,parmar2024one,kang2024distilling} supervise the student at the distribution level by minimizing an approximate reverse KL divergence~\cite{franceschi2023generative,luo2023diffinstruct} or through adversarial learning~\cite{goodfellow2014generative,isola2017image}. 
This enables architectural flexibility between teacher and student, and provides stronger global supervision. Our method builds upon DMD to distill a bidirectional teacher into an efficient student. However, DMD loss alone is insufficient for our task, which motivates us to introduce new modules that leverage input pose sequences.

\paragraph{Video Generation Acceleration.} 
Accelerating video diffusion models~\cite{ho2022video, blattmann2023stable, Peebles2022DiT} is critical due to their high computational cost. Common strategies~\cite{lin2025diffusion, zhai2024motion, lin2024animatedifflightning} include reducing denoising steps using model distillation~\cite{yin2024onestepdmd,kang2024distilling,luhman2021knowledge} or accelerating attention computation. 
This includes causal attention, inherent in autoregressive models~\cite{alonso2024diffusion,jin2024pyramidal,chen2025diffusion}, where the current frame only attends to the past frames. This also enables streaming video generation and long video synthesis~\cite{yin2025causvid}. Other methods use attention sparsity~\cite{child2019generating}, for instance, by dynamically identifying spatiotemporal patterns~\cite{xi2025sparse}. However, existing methods do not fully exploit domain-specific cues, such as those available in co-speech video generation. In contrast, our method leverages input human pose information to design sparse attention and distillation loss. Our method outperforms generic video distillation methods by a large margin.

%% file: sections/3_method.tex
\section{Method}
\lblsec{method}
\input{figText/pipeline}
We introduce a real-time, audio-to-video generation model that creates a co-speech video from an audio clip and a speaker's reference image. Our key idea is to distill a slow teacher video model into a fast student model using input-aware sparse attention and distillation. Below, we first describe our co-speech video generation pipeline in \refsec{cospeech}. Then, we introduce a sparse attention mechanism conditioned on the input human poses in \refsec{attention}, which dramatically reduces computation costs. Finally, \refsec{distillation} describes our distillation objective, which further increases the quality of crucial regions such as faces and hands. 
\input{figText/method}
\subsection{Co-Speech Gesture Video Generation }
\label{sec:cospeech}
As shown in \reffig{pipeline}, we adopt a two-stage pipeline for co-speech gesture video generation, which consists of generating motion sequences from audio and synthesizing the corresponding gesture videos. For the first stage, we build on the EMAGE framework~\cite{liu2023emage}, which converts speech audio and a reference image into temporally aligned upper body motion sequences, including facial expressions, hand gestures, and body movements. These motion sequences are represented as dense pose keypoints to capture structural motion information. In the second stage, we leverage input-conditioned video diffusion models~\cite{zhang2024mimicmotion} as our teacher model, which takes a reference image and a motion sequence to synthesize temporally coherent and identity-consistent videos. This pipeline forms the basis for our proposed acceleration method,  as described in the following sections.

\subsection{Input-Aware Sparse Attention}
\label{sec:attention}
We operate on a video sequence consisting of $T$ frames, where each frame is represented by $N$ tokens. Let $t_q, t_k \in \{1, \ldots, T\}$ be the indices for a query and a key frame, and $i, j \in \{1, \ldots, N\}$ be the respective token indices within these frames. To eliminate irrelevant interactions and accelerate attention computation, we employ an attention mask $\mathbf{M}(t_q, i, t_k, j) \in \{0, -\infty\}$, where $\mathbf{M}(t_q, i, t_k, j) =0$ indicates that token $i$ in frame $t_q$ is permitted to attend to token $j$ in frame $t_k$. Conversely,  $\mathbf{M}(t_q, i, t_k, j) =-\infty$ denotes that attention between the token pair is suppressed.  We next introduce global and local attention masking mechanisms designed to specify which frames and which regions the model should attend to, respectively. 

\paragraph{Input-Aware Global Attention Masking.} 
To ensure our temporal attention mechanism focuses on the relevant past information and adheres to causality, we introduce an input-aware global attention mask $\mathbf{M}_{\text{global}}$. This mask guides the attention computation by identifying a pertinent subset of historical frames for each current frame $t_q$.

For each frame $t \in \{1, \ldots, T\}$, we represent its pose with $B$ upper-body keypoints, denoted as $\mathbf{P}_t \in \mathbb{R}^{B \times 2}$. To quantify pose similarity between the current frame $t_q$ and a historical frame $t_k < t_q$, we consider a global transformation matrix $\tau$ applied to $\mathbf{P}_{t_k}$ to compensate for the subject's global movement. The similarity $S(t_q, t_k)$ is then computed as the minimum alignment error, defined by: 

\begin{equation}
S(t_q, t_k) = \min_{\tau} \| \mathbf{P}_{t_q} - \tau(\mathbf{P}_{t_k}) \|_2, 
\end{equation} 
where $\tau$ is a rigid transformation matrix. 
For each current frame $t_q$, we select the top-$K$ most similar historical frames to construct the attention mask accordingly. Let $\mathcal{S}_{t_q} = \{ S(t_q, t_k) \mid t_k < t_q \}$ be the set of similarity scores with previous frames, and $\mathcal{S}_{t_q}^{(K)}$ be the $K$-th smallest value in $\mathcal{S}_{t_q}$.

We define input-aware global attention mask $\mathbf{M}_{\text{global}}(t_q, t_k)$ as:

\begin{equation}
\mathbf{M}_{\text{global}}(t_q, i, t_k, j) =
\begin{cases}
0, & \text{if } S(t_q, t_k) \leq \mathcal{S}_{t_q}^{(K)} \\
-\infty, & \text{otherwise.}
\end{cases}
\label{eq:global_mask_definition}
\end{equation}

When applied to the attention logits prior to the softmax operation, this mask ensures that only the top-$K$ most similar historical frames contribute to the attention computation for frame $t_q$. We call this the global attention mask because its value is constant for all token pairs $(i,j)$ between any two frames $t_q$ and $t_k$, as it depends only on these frame indices.

\paragraph{Input-Aware Local Attention Masking.}
\label{sec:pose_aware_masking} 
To further enforce local consistency and focus temporal attention on relevant body parts in human subjects, we introduce an input-aware mask $\mathbf{M}_{\text{local}}$. The mask partitions tokens into coherent local regions, as defined by keypoint locations estimated via the rigid moving least squares transformation~\cite{schaefer2006mls}. This formulation supports dense attention within each frame while constraining inter-frame attention to correspondences across homologous local regions. Let $\mathcal{R} = \{\text{faces, hands, arms, bodies, shoulders}\}$ be the set of local regions. Each region $r \in \mathcal{R}$ corresponds to a fixed subset of token indices $\mathcal{I}_r \subseteq \{1, \ldots, N\}$. We then define the local attention mask $\mathbf{M_{\text{local}}}(t_q, i, t_k, j)$ as:

\begin{equation}
\mathbf{M_{\text{local}}}(t_q, i, t_k, j) =
\begin{cases}
0, & \text{if } t_q = t_k \\[6pt]
0, & \text{if } t_q \neq t_k \text{ and } \exists r \in \mathcal{R} \text{ s.t  } i \in \mathcal{I}_r \text{ and } j \in \mathcal{I}_r \\[6pt]
-\infty, & \text{otherwise}.
\end{cases}
\label{eq:local_mask_definition} 
\end{equation}

This formulation ensures that inter-frame attention focuses on locally corresponding areas (e.g., face regions attending to other face regions), while intra-frame attention remains dense to capture local context.

\paragraph{Input-Aware Sparse Attention Mechanism.}
\label{sec:sparse_attention_mechanism}
We adopt $\mathbf{M}$ as our final mask for attention, where $\mathbf{M} = \mathbf{M}_{\text{global}} + \mathbf{M}_{\text{local}}$.  This design enables the attention mechanism to adapt dynamically to the input signals, focusing on critical regions (e.g., faces and hands) while minimizing redundant computation in static areas, such as the background.  

To ensure efficient execution on modern hardware, attention is computed over blocks rather than individual $1 \times 1$ tokens. Specifically, we apply a pooling operation with a block size of $128 \times 128$~\cite{guo2024blocksparse} to construct the sparse attention mask by Flashinfer~\cite{Ye2025FlashInfer_MLSys}, which substantially reduces the number of attention queries while preserving fine-grained structure. 

\subsection{Input-Aware Model Distillation}
\label{sec:distillation}
Now we distill the original multi-step teacher video diffusion model with full attention into an efficient student model with input-aware sparse attention. This enables real-time video generation while maintaining perceptual quality.
\paragraph{DMD Loss. } Following variational score distillation~\cite{wang2023prolificdreamer} and distribution matching distillation (DMD)~\cite{yin2024onestepdmd,yin2024improveddmd}, the student model $G_{\theta}$ is trained to match the latent distributions of the teacher in fewer timesteps. Specifically, the student minimizes the reverse Kullback-Leibler (KL) divergence between the student's data distribution $p_{\text{gen}}(\x_t)$ and the teacher's data distribution $p_{\text{data}}(\x_t)$ at randomly sampled timesteps $t$:

\begin{align}
\nabla_{\theta} \mathcal{L}_{\mathrm{DMD}} &\triangleq \mathbb{E}_{t} \left[ \nabla_{\theta} \mathrm{KL} \big( p_{\mathrm{gen}, t} \| p_{\mathrm{data}, t} \big) \right] \\
&\approx - \mathbb{E}_{t} \left[ \big( s_{\mathrm{data}}(\x_t, t) - s_{\mathrm{gen}}(\x_t, t) \big) \frac{\partial G_{\theta}}{\partial \theta} \right]. \nonumber
\end{align} 
where \( s_{\mathrm{data}}(\x_t, t) \) and \( s_{\mathrm{gen}}(\x_t, t) \) denote the score functions of the teacher and student models at timestep \( t \) for the input \( \x_t \), respectively. This formulation encourages the student to approximate the teacher's denoising behavior with fewer steps, achieving a balance between efficiency and fidelity.

\paragraph{Input-Aware Distillation Loss.}
We compute an input-aware loss over all regions by applying masks $\mathbf{m}_r$ to the full ground-truth image $\x$ and the generated image $\hat{\x}$. These masks are specific to each image, constructed from its pose information to delineate the same region type $r$. The unified loss is formulated as follows:

\begin{equation}
\mathcal{L}_{\text{region}} = \sum_{r\in \mathcal{R}} \lambda_r \, \mathcal{L}_r\big(\mathbf{m}_r \odot \x, \mathbf{m}_r \odot \hat{\x}\big),
\label{eq:region_loss_mask}
\end{equation} 
where $\odot$ denotes the Hadamard product (element-wise multiplication), $\lambda_r$ balances the importance of each region, and $\mathcal{L}_r(\cdot, \cdot)$ denotes the reconstruction loss for each region. 
Specifically, for the face region, we extract feature embeddings using a pre-trained ArcFace network~\cite{deng2019arcface} and compute L2 distance between the two, while for other regions, LPIPS metric~\cite{zhang2018perceptual} is used. This approach leverages specialized metrics for distinct regions. Face-specific and gesture-specific losses have been used in prior works in video generation~\cite{richardson2021encoding,zhang2024mimicmotion}. Here, we adopt them for the video distillation task.

\paragraph{Final Objective.} Combining the distribution matching distillation loss and the input-aware distillation loss, the overall training objective for the student model \( G_{\theta} \) is

\begin{equation}
G_{\theta}^* = \arg\min_{G_{\theta}} \quad \mathcal{L}_{\mathrm{DMD}}(G_{\theta}) + \lambda_{\text{region}} \, \mathcal{L}_{\text{region}}(G_{\theta}),
\label{eq:final_loss}
\end{equation}
where \( \lambda_{\text{region}} \) is a scalar hyperparameter controlling the trade-off between matching the teacher's latent distributions and preserving semantic regional consistency.

%% file: figText/pipeline.tex
\begin{figure}[t]
    \centering
    \includegraphics[width=1\columnwidth]{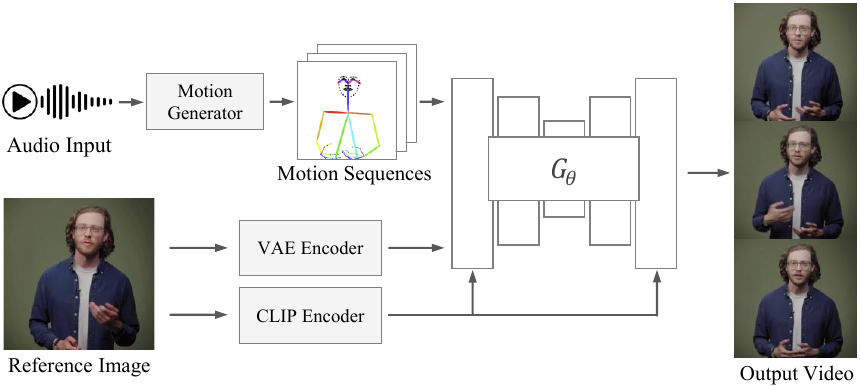}
    \caption{
    \textbf{Our two-stage co-speech video generation pipeline.} In Stage 1, an audio input and a reference image are fed into an audio-to-motion generator~\cite{liu2023emage}  to produce motion sequences represented by dense pose keypoints. In Stage 2, these motion sequences are fed into our efficient student video generation network $G_{\theta}$, The network is conditioned on features from the reference image, which are separately encoded by the VAE encoder~\cite{rombach2022high} and CLIP encoder~\cite{radford2021learning}, to synthesize the final video. In our work, we focus on accelerating the video generation, significantly speeding up the teacher model~\cite{zhang2024mimicmotion}.}
    \vspace{-1.8pt}
    \lblfig{pipeline}
\end{figure}

%% file: figText/method.tex
\begin{figure*}[t]
    \centering
    \includegraphics[width=1\textwidth]{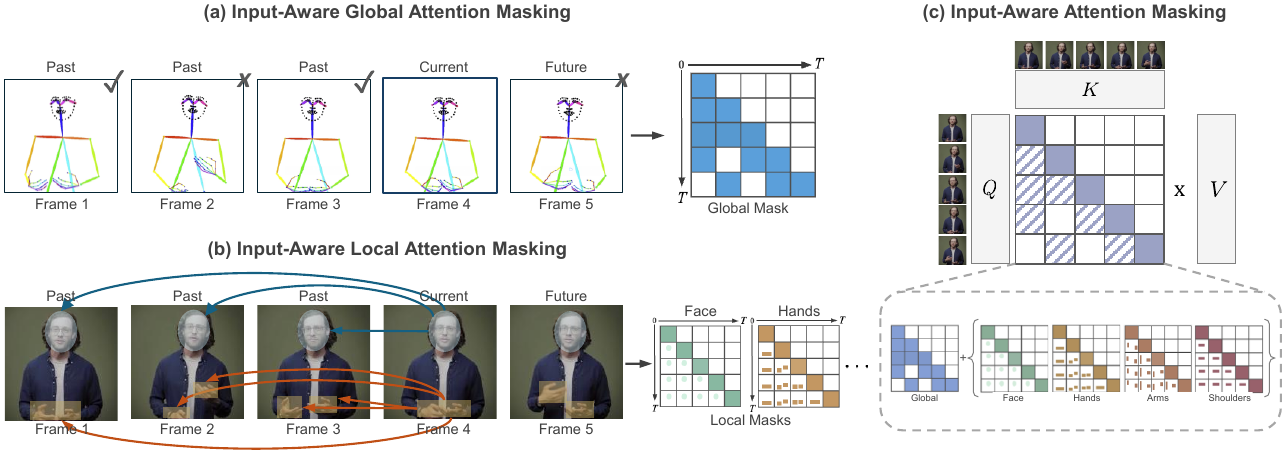}
    \caption{\textbf{Input-Aware Sparse Attention.} 
    Our attention mechanism selectively focuses on tokens within salient body regions and their corresponding areas in temporally relevant frames.
    (a) We first apply global masking, which restricts attention to the $K$ most similar past frames based on pose similarity.
    (b) Then local masking limits inter-frame attention to matched regions (e.g., face, hands) to enhance temporal coherence.
    (c) Our input-aware attention masking integrates both global and local masks to form an efficient and structured sparse attention pattern.}
    \label{fig:attention_masking}
\end{figure*}

%% file: sections/4_result.tex
\section{Experiments}
\label{sec:experiment}
\subsection{Experimental Setups}
\input{figText/figure_page}
\input{tblText/baseline}
\paragraph{Datasets.} We evaluate our method on two datasets: the public TalkShow dataset~\cite{yi2023generating}, which contains four speakers with varied backgrounds and irregular camera movements, and our curated YouTube Talking Video dataset. Our YouTube dataset features 45 speakers with diverse appearances in half-body and headshot views, set against clean backgrounds with static cameras.

To extract human poses, we use DWPose~\cite{yang2023effective} and apply a strict filtering pipeline to ensure high-quality pose information. We retain only frames where (1) facial keypoint confidence exceeds 0.9, (2) upper-body keypoint confidence exceeds 0.8, and (3) upper-body keypoint visibility is at least 90\%. This filtering process yields reliable upper-body and frontal-face cues for training.

\paragraph{Metrics.}
We evaluate our method using a comprehensive set of metrics, including image and video fidelity (FID~\cite{heusel2017gans}, FVD~\cite{unterthiner2019fvd}, E-FID~\cite{deng2019efid}), image quality (PSNR, SSIM~\cite{Wang2004Image}), audio-lip synchronization (Sync-C, Sync-D)~\cite{prajwal2020lip}, hand motion quality and diversity(HKC, HKD)~\cite{lin2024cyberhost}, and inference efficiency (FPS). Specifically, Sync-C and Sync-D measure the alignment between speech and lip movements, while E-FID evaluates the alignment of emotional expressions between the generated and real videos. HKC and HKD evaluate the average detection confidence and variance of the hand keypoints in the generated sequence. The reported FPS is measured on an NVIDIA H100 GPU, excluding I/O latency, to accurately reflect the model's performance.

\subsection{Implementation Details}
Similar to the teacher model, we train our model across speakers with diverse visual and motion characteristics. The training procedure consists of two stages. First, we fine-tune the teacher model by incorporating our input-aware sparse attention, utilizing 4 NVIDIA H100 GPUs for approximately 6 hours. Second, we distill the student model from the adapted teacher, which takes about 10 hours on 8 NVIDIA H100 GPUs.

\subsection{Comparisons}
We compare our method with recent open-source methods in both audio-driven and pose-driven video generation settings. For audio-driven baselines MMDiffusion~\cite{ruan2022mmdiffusion} and S2G-MDD~\cite{he2024co}, which are trained per speaker, we perform our evaluation on the same TalkShow~\cite{yi2023generating} dataset to ensure a fair comparison.  For pose-driven baselines, we compare against  AnimateAnyone~\cite{hu2023animateanyone}, EchoMimicV2~\cite{meng2024echomimicv2},  and MimicMotion~\cite{zhang2024mimicmotion} on our YouTube Talking dataset. To evaluate the complete co-speech pipeline, we use our audio-to-motion module to generate pose sequences as input to these methods. EchoMimicV2~\cite{meng2024echomimicv2} accepts audio as input, its video generation strictly relies on externally provided pose sequences rather than audio-driven motion prediction. Therefore, we categorize it as a pose-driven method. 

\paragraph{Quantitative Results.}
As detailed in Table~\ref{tab:baseline}, our method achieves approximately $3\times$ faster inference compared to existing audio-driven and pose-driven baselines.
In addition to speed, our approach produces higher-quality and more realistic results. Compared with audio-driven methods, our model not only maintains high generation quality but also substantially improves Sync-C and HKC. More specifically, the lip synchronization confidence significantly improves from 4.36 to 7.26. Additionally, compared to S2G-MDD, our method improves HKC from 0.956 to 0.968 on the test set.

Compared to pose-driven methods, our approach outperforms others in both lip synchronization and overall motion quality. Remarkably, our student model, when compared with its teacher model MimicMotion~\cite{zhang2024mimicmotion}, not only achieves an impressive $13.1\times$ inference acceleration without sacrificing generation quality, but also further enhances motion and synchronization quality. Specifically, our model substantially improves HKC from 0.928 to 0.948 and Sync-C from 4.56 to 7.28, demonstrating enhanced hand motion confidence and lip synchronization.

\input{figText/ablation}

\input{tblText/ablation}

\paragraph{Qualitative Results.}
As shown in \reffig{speech_baseline}, our method demonstrates clear improvements over existing audio-driven methods in lip-audio synchronization, expressive hand gestures, and overall visual quality.  Specifically, our generated videos exhibit better lip-audio synchronization, where the lip movements align more naturally with the speech content. Additionally, the overall visual quality is significantly higher, producing sharper and more realistic appearances compared to the blurry or less realistic results from the baselines.
\reffig{pose_baseline} shows that our method generates more natural lip and hand animations than pose-driven baselines. Existing pose-driven methods often produce stiff or unnatural movements in these critical regions. In contrast, our model maintains high fidelity and realism, with lifelike facial and hand animations, while achieving faster inference.
\subsection{Ablation Study}
We conduct ablation studies on the YouTube Talking Video dataset.

\paragraph{Ablation Study on Model Components.} We conduct an ablation study to analyze the contribution of each component, with results in Table~\ref{tab:ablation} and Figure~\ref{fig:ablation}. 
Leveraging the teacher model as the baseline, we observe strong motion and lip synchronization performance after being fine-tuned on the co-speech dataset. However, the inference speed remains a bottleneck at 1.93 FPS. While input-aware global attention and input-aware local attention maintain generation quality without degradation, the speed improvements are limited. Direct application of DMD distillation achieves a significant speedup but results in quality degradation, particularly noticeable artifacts on faces and hands.  By incorporating our input-aware distillation, the model achieves real-time performance at 25.31 FPS while maintaining generation quality comparable to the finetuned Teacher model. 

\paragraph{Ablation Study on Efficiency.}
We present a detailed breakdown of the inference time across different architectural variants by decomposing the total runtime into four components: Attention, Linear, Norm, and Others in Figure~\ref{fig:efficiency}. The baseline Teacher model requires 103.6 seconds to process an 8-second video. By introducing Global Attention, we reduce the time to 60.9 seconds, mainly due to reductions in attention and linear computations. Local Attention further decreases the time to 45.2 seconds. Finally, applying distillation reduces the runtime to 7.9 seconds, a $13.1\times$ speedup compared to the Teacher model, primarily attributed to the substantial reduction in attention costs. This demonstrates the effectiveness of our sparse attention strategy in enabling real-time co-speech video generation.

\input{figText/efficiency}
\subsection{Results}
\paragraph{Additional Qualitative Results.} 
In \reffig{our}, we present additional qualitative examples generated by our method to illustrate its visual quality and expressiveness. Conditioned on a single static reference image and an input audio clip, our model synthesizes temporally coherent video sequences with high appearance fidelity and natural motion. The results demonstrate lifelike facial expressions, smooth upper body movements, and precise synchronization of the audio lip, even under variations in speaker identity, pose, and expression. These examples highlight the robustness of our method in capturing fine-grained motion dynamics while preserving speaker identity across frames. In addition, they complement the quantitative evaluations by providing intuitive evidence that our method can generate visually convincing talking videos in real time, underscoring its potential for applications in human–computer interaction and digital content creation.

\paragraph{Long Video Generation.}
We adopt a progressive strategy for long video generation that explicitly enforces temporal smoothness across extended sequences. At each denoising step, the video is partitioned into consecutive segments, and each segment is generated independently by conditioning the trained model on a fixed reference image together with its corresponding sub-sequence of poses. To guarantee consistency between adjacent segments, the overlapped frames are progressively fused according to their temporal indices. This progressive design not only mitigates the accumulation of temporal artifacts but also preserves global coherence throughout the sequence, enabling our method to stably generate videos of several minutes in length without noticeable degradation in visual quality or motion dynamics.

\paragraph{User Preference Study.}
We conduct a user preference study on Amazon Mechanical Turk (AMT) with 30 participants to evaluate the perceptual quality of our generated videos. Specifically, each participant is asked to perform three separate pairwise comparisons over 72 videos: (1) our method vs. audio-driven generation, (2) our method vs. pose-driven generation, and (3) our method vs. ground truth (GT). In each comparison, participants are presented with videos generated from identical reference images and audio inputs, displayed in randomized order without revealing the underlying method. For each trial, users are asked to indicate their preference in terms of lip synchronization, motion realism, and overall video quality. As reported in Table~\ref{tab:user_study}, our method is favored by users in comparison to baselines, and its performance is comparable to the ground truth.

%% file: figText/figure_page.tex
\begin{figure*}[htbp] 
\centering 
\includegraphics[width=0.88\linewidth]{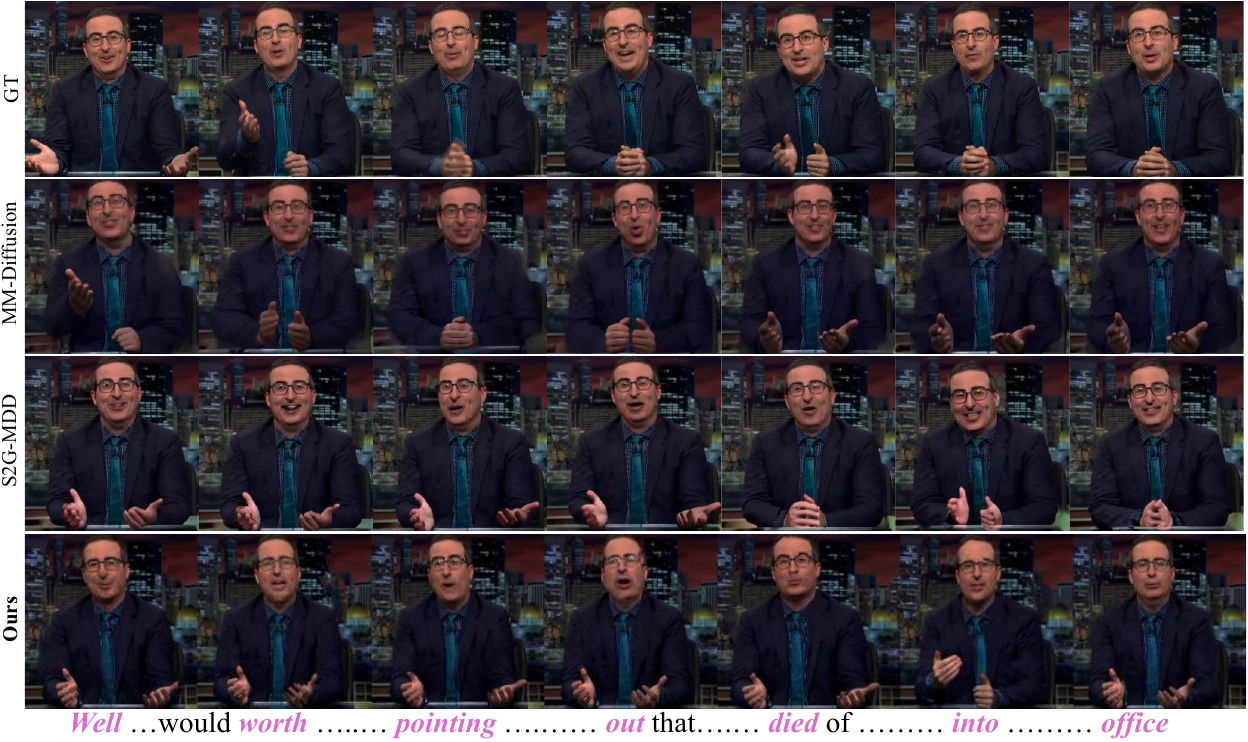} 
\vspace{-12pt} 
\caption{\small \textbf{Qualitative comparison of audio-driven methods.} We show body animation results conditioned on the same input audio. Our method not only achieves accurate lip synchronization but also generates clearer hand gestures.} \lblfig{speech_baseline} 
\end{figure*} 

\begin{figure*}[htbp] 
\centering 
\includegraphics[width=0.88\linewidth]{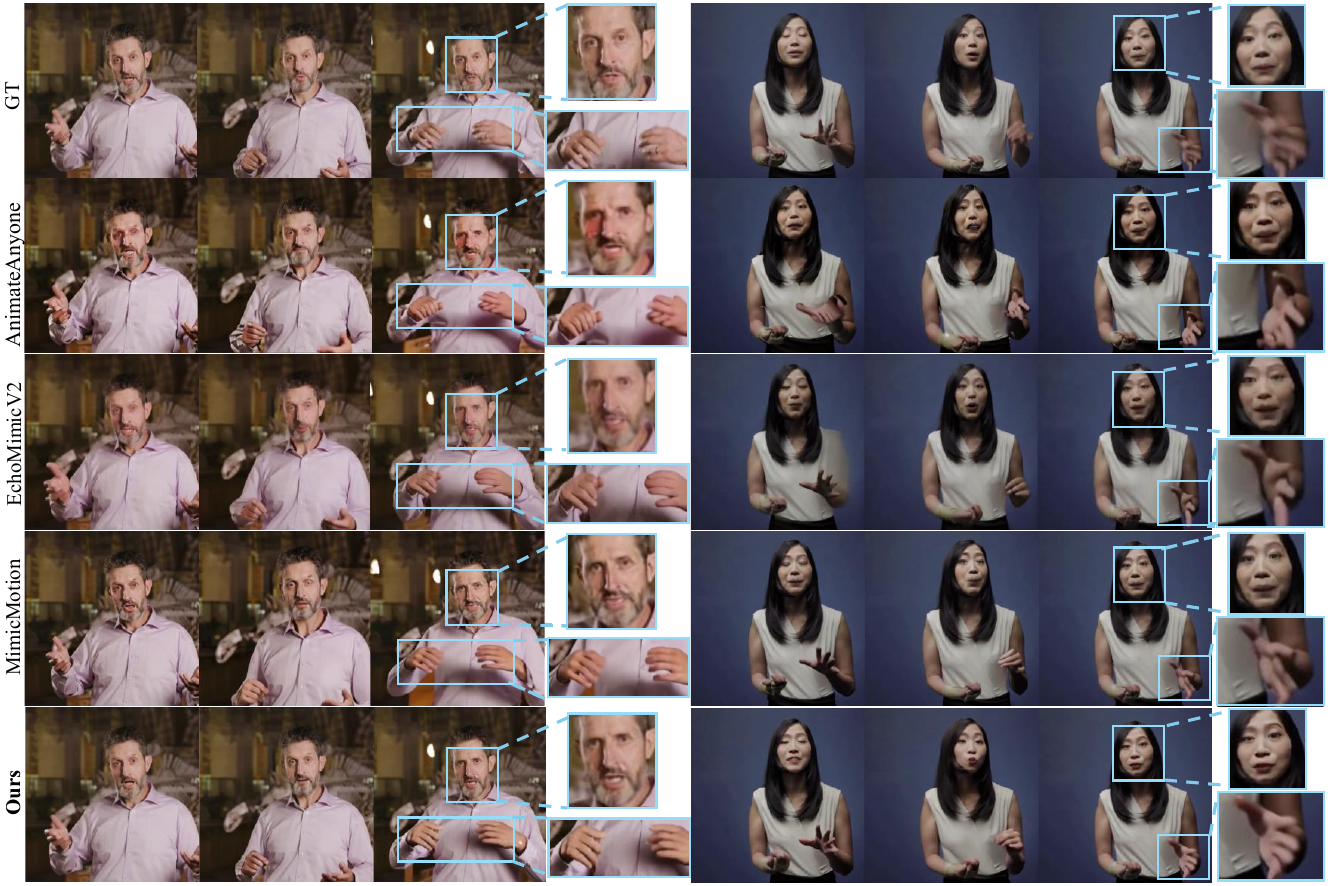} 
\vspace{-12pt} 
\caption{\small \textbf{Qualitative comparison of pose-driven methods.} All methods are conditioned on the same motion sequence and reference image. Our approach produces more realistic faces, hands, and body movements.} 
\lblfig{pose_baseline} 
\end{figure*}

\begin{figure*}[htbp]
    \centering
    \includegraphics[width=0.85\linewidth]{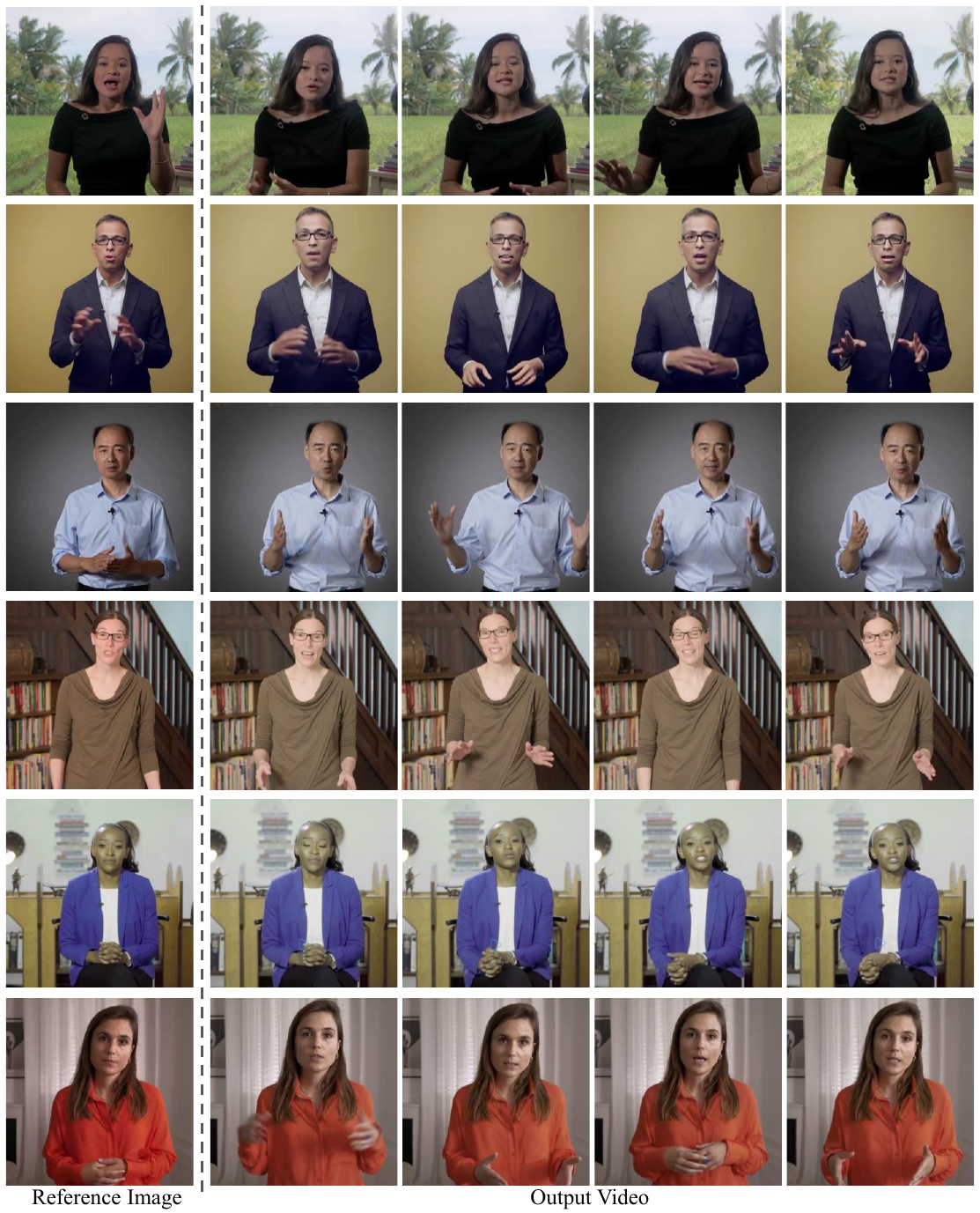}

    \caption{\small \textbf{Additional Qualitative Results.} This figure presents a selection of further examples generated by our method. Given only a single static reference image and an input audio clip, our model effectively synthesizes highly realistic and expressive video outputs. These results visually demonstrate its capability to produce natural facial expressions, fluid body movements, and accurate lip synchronization in real time.}

        \lblfig{our}
\end{figure*}

%% file: tblText/baseline.tex
\begin{table*}[t]
\small
\tabcolsep 4.5pt
\centering
\caption{\textbf{Quantitative comparison with state-of-the-art methods.} 
We evaluate audio-driven methods on the TalkShow dataset~\cite{yi2023generating} and pose-driven methods on the YouTube Talking Video dataset. 
Metrics include image and video fidelity, audio-lip synchronization and hand motion accuracy.
\textbf{Bold} indicates the best result within each group of audio- or pose-driven methods.}
\label{tab:baseline}

\renewcommand{\arraystretch}{1.15}
\begin{tabular}{lccccccccccc}
\toprule
Method & FPS $\uparrow$ & FID $\downarrow$ & FVD $\downarrow$ & SSIM $\uparrow$ & PSNR $\uparrow$ & HKC $\uparrow$ & HKD $\uparrow$ & Sync-C $\uparrow$ & Sync-D $\downarrow$ & E-FID $\downarrow$ \\
\midrule
\multicolumn{11}{l}{\textbf{Audio-Driven}} \\
\midrule
MMDiffusion~\cite{ruan2022mmdiffusion}   & 2.82  & 81.25 & 985.06 & 0.549 & 17.62 & 0.946 & 23.24 & 3.44 & 8.02 & 4.71 \\
S2G-MDD~\cite{he2024co}                  & 6.89  & 68.11 & 883.48 & 0.552 & 19.00 & 0.956 & 23.44 & 4.36 & 7.59 & 3.39 \\
\textbf{Ours}                            & \textbf{25.31} & \textbf{66.81} & \textbf{823.58} & \textbf{0.596} & \textbf{19.40} & \textbf{0.968} & \textbf{24.26} & \textbf{7.26} & \textbf{6.98} & \textbf{3.16} \\

\midrule
\multicolumn{11}{l}{\textbf{Pose-Driven}} \\
\midrule
AnimateAnyone~\cite{hu2023animateanyone} & 2.07  & 61.23 & 759.83 & 0.752 & 20.94 & 0.921 & 24.63 & 2.00 & 9.38 & 5.08 \\
EchoMimicv2~\cite{meng2024echomimicv2}   & 8.92  & 57.98 & 611.65 & 0.800 & 22.02 & 0.939 & 24.83 & 7.19 & 7.02 & \textbf{2.97} \\
MimicMotion~\cite{zhang2024mimicmotion}  & 1.93  & \textbf{56.32} & 628.03 & 0.823 & \textbf{23.89} & 0.928 & 24.82 & 4.56 & 7.34 & 3.38 \\
\textbf{Ours}                            & \textbf{25.31} & 57.01 & \textbf{610.34} & \textbf{0.829} & 23.42 & \textbf{0.948} & \textbf{24.83} & \textbf{7.28} & \textbf{6.99} & 3.01 \\

\bottomrule
\end{tabular}
\end{table*}

%% file: figText/ablation.tex
\begin{figure}[t]
    \centering
    \includegraphics[width=0.95\columnwidth]{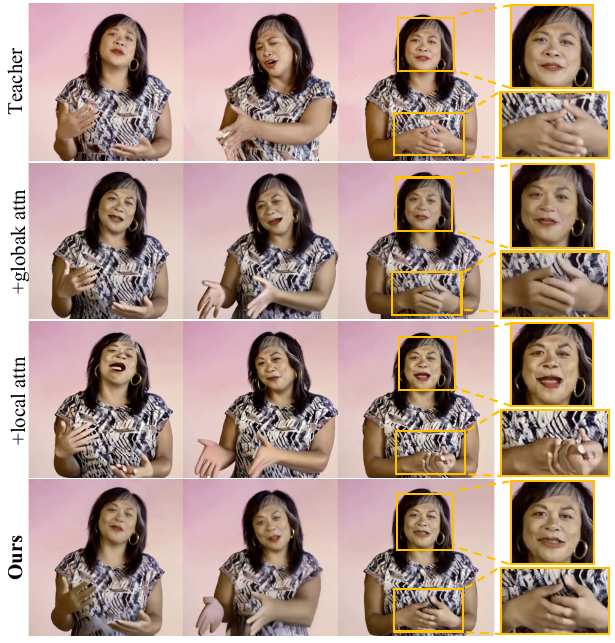}
    \caption{\small \textbf{Qualitative ablation on model components.} While the combination of global attention and causal attention improves efficiency, it introduces degradation in the hand and mouth regions. Our input-aware distillation not only restores visual fidelity and temporal consistency, but also achieves real-time performance with natural and synchronized motion.}
    \lblfig{ablation}
\end{figure}

%% file: tblText/ablation.tex
\begin{table}[t]
\small
\centering
\caption{\small \textbf{Quantitative ablation on model components.} We analyze the effect of progressively incorporating architectural components. Introducing input-aware attention and distillation substantially accelerate inference but degrade quality. In contrast, our input-aware distillation achieves improved motion realism accuracy while maintaining real-time performance.}
\label{tab:ablation}
\renewcommand{\arraystretch}{1.2}
\setlength{\tabcolsep}{4pt}
\begin{tabular}{lccccc}
\toprule
Method& FPS $\uparrow$ & SSIM $\uparrow$ & PSNR $\uparrow$ & HKC $\uparrow$ & Sync-C $\uparrow$  \\
\midrule
Teacher                  & 1.93 & 0.823 & 23.89 & 0.928 & 4.56 \\
\midrule
+finetuned     & 1.93 & \textbf{0.835} & \textbf{24.01} & 0.945 & 7.15 \\
+global attention      & 3.28 & 0.834 & 23.89 & 0.943 & 7.14 \\
+local attention     & 4.42 & 0.831 & 23.89 & 0.943 & 7.14 \\
+$\mathcal{L}_{\mathrm{DMD}}$ & 25.31 & 0.826 & 22.99 & 0.941 & 7.08\\
\midrule
+\textbf{$\mathcal{L}_{\mathrm{Region}}$ (Ours)} & \textbf{25.31} & 0.829 & 23.87 & \textbf{0.948} & \textbf{7.28} \\
\bottomrule
\end{tabular}
\end{table}

%% file: figText/efficiency.tex
\begin{figure}[t]
 \centering
\includegraphics[width=0.98\columnwidth]{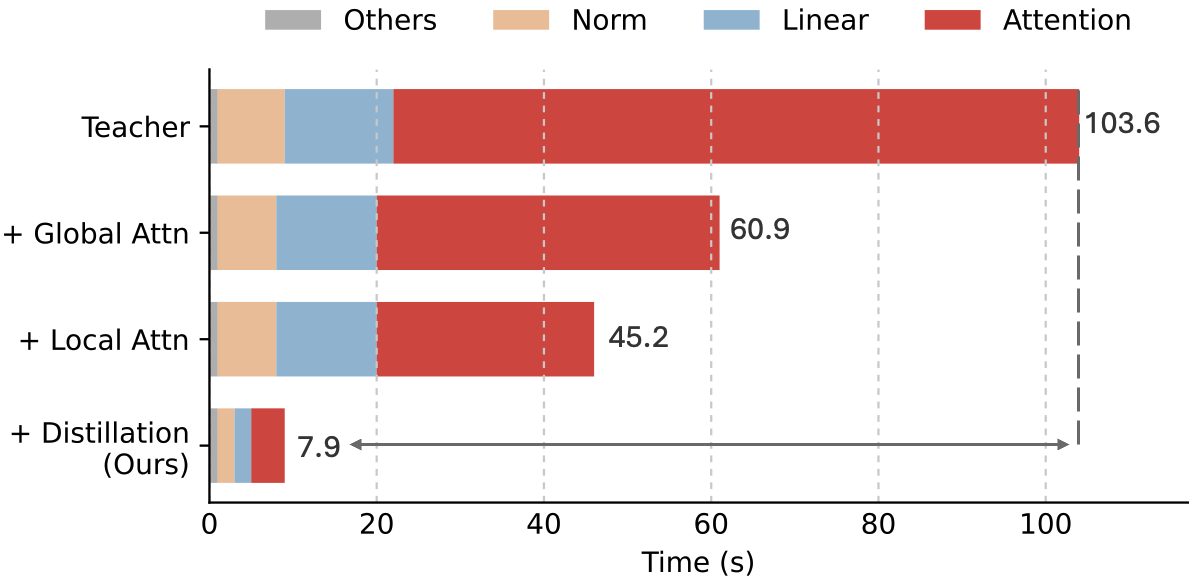}
\caption{\small \textbf{Ablation on Efficiency.} 
We compare the inference time with the incorporation of input-aware attention and distillation. Our method significantly reduces computation while preserving quality, enabling real-time performance.}
\vspace{-1pt}
\lblfig{efficiency}
\end{figure}

%% file: sections/5_conclusion.tex
\section{Discussion and Limitations}
\input{tblText/user_study}

\label{sec:discussion}
\paragraph{Limitations.} 
Our current method still has several limitations. First, we cannot handle videos with dynamic backgrounds or complex scene motion, as our input-aware sparse attention assumes a relatively static camera and background. This limits the applicability of our approach to studio-like or controlled environments.
Second, while our method improves lip synchronization and hand motion realism, it still struggles with subtle gestures, such as finger articulation, as shown in Figure~\ref{fig:failure}. Third, our current system is trained and evaluated primarily on monolingual English-speaking datasets. Extending to multilingual or code-switching scenarios may require additional modeling of phoneme-to-gesture dynamics across languages.
Lastly, although our model runs in real-time, it still requires a relatively powerful GPU for inference, which may restrict its deployment on edge devices or mobile platforms.

\paragraph{Ethics Considerations.} 
Our research advances generative models for human video synthesis. We recognize that this technology may also be misused for malicious purposes such as spreading misinformation, damaging reputations, or producing deceptive content. Prior studies~\cite{wang2019cnngenerated, cozzolino2021reveal} on deepfake detection have shown that generated videos often contain detectable artifacts, underscoring the importance of video forensics in mitigating these risks. However, current forensic techniques remain limited, and the continued advancement of generative models necessitates parallel progress in detection to ensure reliability.
 
\paragraph{Conclusion.} 
In this work, we introduced a new conditional video distillation framework specifically designed for fast and high-quality co-speech video generation. Our core approach leverages pose information as a natural and effective conditioning signal, guiding both the attention and the supervision mechanisms of our student model. To achieve this, we proposed an input-aware sparse attention module that intelligently focuses computation on dynamically important regions across frames, alongside an input-aware distillation loss which prioritizes visual fidelity in critical  areas like the face and hands. Collectively, these components enabled our distilled model to achieve real-time inference speeds, all while not only preserving but significantly enhancing motion realism and lip synchronization quality compared to the teacher model.
\input{figText/failure}
\begin{acks}
 We would like to thank Kangle Deng, Muyang Li, Nupur Kumari, Sheng-Yu Wang, Maxwell Jones, Gaurav Parmar for their insightful feedback and input that contributed to the finished work. The project is partly supported by Ping An Research.
\end{acks}

%% file: tblText/user_study.tex
\begin{table}[t]
\small
\centering
\caption{\textbf{User Preference Study.} Results of a user study evaluating the perceptual quality of video generation. Participants were asked to compare our method against baseline approaches along three aspects: Lip Synchronization, Motion Realism, and Video Quality. The table reports preference rates (\%) across all participants for each method.}
\label{tab:user_study}
\renewcommand{\arraystretch}{1.05}
\begin{tabular}{lccc}
\toprule
\multirow{2}{*}{Method} & \multicolumn{3}{c}{User Preference} \\
\cmidrule(lr){2-4}
 & Lip Sync & Motion Realism & Video Quality \\
\midrule
MMDiffusion & 2.3\% & 11.1\% & 2.2\% \\
S2G-MDD      & 13.2\% & 15.6\% & 6.7\% \\
\textbf{Ours}          & \textbf{84.5\%} & \textbf{73.3\%} & \textbf{91.1\%} \\
\midrule
AnimateAnyone & 8.9\% & 16.7\% & 18.9\% \\
EchoMimicV2                & 31.1\% & 24.4\% & 24.4\% \\
MimicMotion               & 12.2\% & 26.7\% & 27.8\% \\
\textbf{Ours}                          & \textbf{47.8\%} & \textbf{32.2\%} & \textbf{28.9\%} \\
\midrule
Ground Truth     & \textbf{54.5\%} & 46.7\% & \textbf{56.6\%} \\
\textbf{Ours}                          & 45.5\% & \textbf{53.3\%} & 43.4\% \\
\bottomrule
\end{tabular}
\end{table}

%% file: figText/failure.tex
\begin{figure}[t]
    \centering
    \includegraphics[width=1\columnwidth]{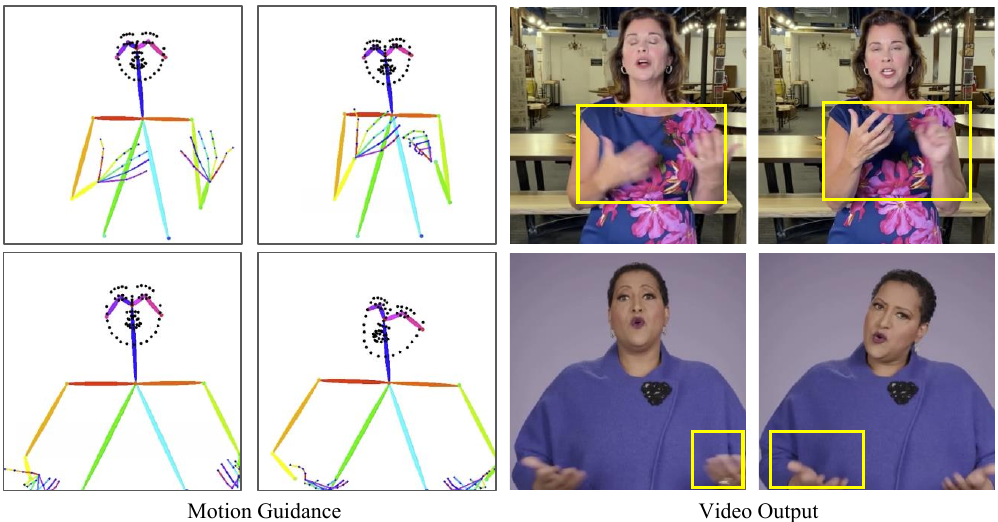}
    \caption{\small \textbf{Limitations.} Given the same motion input (left), we observe that when fingers undergo fine-grained movements, our model tends to produce blurry or distorted hand regions.}
    \lblfig{failure}
\end{figure}